# Large Language Models for Knowledge Graph Embedding: A Survey

Bingchen Liu[1], Yuanyuan Fang[2], Naixing Xu[1], Shihao Hou[3], Xin Li[1*], Qian Li[1*]

[1*] School of Software, Shandong University, Jinan, China.
[2] School of Computer and Artificial Intelligence, Shandong University of Finance and Economics, Jinan, China.
[3] Fujian Key Laboratory of Sensing and Computing for Smart City, School of Informatics, Xiamen University, Xiamen, 361005, China.

*Corresponding author(s). E-mail(s): lx@sdu.edu.cn; feiwangyuzhou@sdu.edu.cn;
Contributing authors: lbcraf2018@126.com; 2896299017@qq.com; naixingxu@mail.sdu.edu.cn; houshihao@stu.xmu.edu.cn;

## Abstract

Large language models (LLMs) have garnered significant attention for their superior performance in many knowledge-driven applications on the world wide web. These models are designed to train hundreds of millions or more parameters on large amounts of text data, enabling them to understand and generate natural language effectively. As the superior performance of LLMs becomes apparent, they are increasingly being applied to knowledge graph embedding (KGE) related tasks to improve the processing results. Traditional KGE representation learning methods map entities and relations into a low-dimensional vector space, enabling the triples in the knowledge graph to satisfy a specific scoring function in the vector space. However, based on the powerful language understanding and semantic modeling capabilities of LLMs, that have recently been invoked to varying degrees in different types of KGE related scenarios such as multi-modal KGE and open KGE according to their task characteristics. In this paper, we investigate a wide range of approaches for performing LLMs-related tasks in different types of KGE scenarios. To better compare the various approaches, we summarize each KGE scenario in a classification. Finally, we discuss the applications in which the methods are mainly used and suggest several forward-looking directions for the development of this new research area.

# 1 Introduction

With the development of society and the advancement of science and technology, knowledge graphs (KGs) have been more and more widely used in applications containing rich semantic information in different fields. At present, KGs have been widely used in various fields such as representation learning [1, 2], bioinformatics networks [3] [4], traffic flow prediction [5] [6], which is proved to have superior performance. Traditional methods for handling the aforementioned tasks typically necessitate learning based on knowledge graph embedding (KGE) representations initially. Subsequently, relevant tasks are carried out on the basis of these precise KGE representations. However, deep learning models in traditional approaches often suffer from performance limitations due to their training scale.

Large Language Models (LLMs) are increasingly being applied to KGE related tasks to enhance their processing because of their superior performance. LLMs can be built upon datasets that are several times larger than those of traditional models. When combined with techniques such as fine - tuning and the use of calling cue words, they can achieve even more remarkable performance in KGE tasks compared to traditional predictive models [7][8][10][9][11]. The aim of this paper is to provide a comprehensive overview and structural categorization of the various task approaches that have been applied to LLMs in different KGE scenarios.

Unlike previous surveys that may have provided fragmented or narrow - focused coverage, this survey aims to offer a truly comprehensive overview of the work related to various KGE scenarios within the context of LLMs. By approaching the topic from the perspective of different methods of calling LLMs, it comprehensively encompasses all aspects of KGE application scenarios and use cases, presenting a more in - depth and holistic analysis that fills the gaps left by previous works. The previous review [117] try to classify LLMs and KGE from the perspective of enhancement methods, or some reviews [118] classify the different stages of KG enhancement and pre-trained language model enhancement based on LLMs. With the continuous progress of research, various emerging application scenarios of KGE are emerging. However, there is still a lack of LLMs classification overview of KGE representation learning in various application scenarios. There have also been a number of surveys about KGE or LLM before. Some articles have been investigated only for KGE aspects. For example, the review written by Liu et al.[12] discusses in detail the current application status, technical methods, practical applications, challenges, and future prospects of KGEs in industrial control system security situational awareness and decision-making. A review by Zhu et al.[13] summarized the KG entity alignment method based on representation learning, proposed a new framework, and compared the performance of different models on diverse datasets in detail, looking forward to future research paths. Some other articles have been investigated only for some applications related to LLMs. For example, a review by Huang et al [14] explores the security and credibility of LLM in industrial applications, analyzed vulnerabilities and limitations, explored integrated and extended verification and validation techniques, and ensured the full lifecycle security of LLMs and its applications. This review by Kumar, P [15] comprehensively explores the applications of LLMs in language modeling, word embedding, and deep learning, analyzes their diverse applications in multiple fields and the limitations of existing methods, and proposes future research directions and potential progress. The above papers describe only one aspect of the investigation. However, our paper comprehensively investigates the novel approach of combining LLMs and KGE. In the meantime, we hope that readers can read the above paper if they are interested, as it will facilitate the reading and understanding of this thesis.

Building upon the research groundwork and surveys presented in the aforementioned related papers, we conduct a more meticulous literature review. Our primary focus lies on the current KGE application scenarios. We categorize and discuss these diverse KGE application scenarios based on the degree of LLM invocations. Through such in-depth descriptions, we achieve a comprehensive exploration of this domain, uncovering nuances and relations that have not been fully examined before. This paper is organized as follows: in Section II, we introduce some basic concepts related to this paper. in Section III, we introduce KGE-related tasks; in Section IV, we discuss the application of LLMs in different KGE scenarios; in Section V, we propose some possible future applications and research directions. in Section VI, we introduce some famous datasets and code resources; finally, we conclude the paper in Section VII.

## 2 Preliminaries

To facilitate the reader's understanding, we introduce relevant basic concepts in this section. We summarize all acronyms used throughout the manuscript in Table II.

### 2.1 Large Language Models (LLMs)

LLMs can be defined as a type of artificial intelligence technology based on deep learning, trained on massive text corpora to generate and understand natural language. Its' basic process involves extracting language patterns from large-scale data, modeling them with neural networks, and generating high-quality language outputs based on user inputs. Its' applications span various fields, including text generation[16], machine translation[17], question answering[18], sentiment analysis[19], and content summarization[20]. Compared to traditional methods, LLMs offer superior contextual understanding, higher output quality, and enhanced transfer learning capabilities.

### 2.2 Heterogeneous Information Network (HIN)

HIN can be defined as a network structure that contains multiple types of nodes and edges, which is capable of modeling rich data types and complex relations in the real world. Compared with traditional homogeneous networks, HINs are able to capture the semantic information and structural features of data in a more comprehensive way [26], which provides new perspectives and methods for data mining and knowledge discovery. The relations between these nodes and edges are not just simple connections, but contain rich semantics, such as purchase, collection, favorite, and so on. Semantic mining can be performed in networks based on different methods to discover deeper patterns and knowledge, which are mainly categorized into four types: edge sampling-based representation strategies [27], path-based representation strategies [28], subgraph-based representation strategies [29], and deep learning-based representation strategies [30].

Research on HINs still faces many challenges and opportunities. Future research can focus on several key directions. One important area is exploring how to construct and analyze more complex HINs more effectively [31]. Another area involves investigating how to better integrate HINs with multimodal data and KGs [32]. Additionally, developing more powerful tools and algorithms for analyzing HINs can also be a critical area of focus. The practical applications of HINs in the fields of business [33], cybersecurity [34], and medicine [35] are also important directions for future research.

## 3 KGE-related Tasks

With the development of society and the progress of science and technology, graph data has been more and more widely used in various fields, such as social networks[36][37], academic networks[38], biological networks[39], etc. Graph data in various domains often consists of multiple types of edges and nodes, which contains rich semantic information, and puts high demands on accurate graph data information representation in various types of downstream applications in different domains. Knowledge graph is a graph-based structured semantic knowledge base that stores graph data of various domains in the form of (entity, relation, entity) ternary structure. KG can be regarded as a special kind of HIN in which entities and relations have clear semantic definitions. Meanwhile, the research methods and techniques of HIN can also be applied in the construction and analysis of KG.

KGE transforms entities and relations in KGs into embedded representations mapped to a low-dimensional continuous vector space. KGE can adequately capture the complex graph data structure and rich semantic information of various domains, and facilitates the management of the huge amount of graph data in different domains, use and maintenance. At present, in various fields such as medicine [40][41], news media [42][43], etc. KGE has been widely used in related applications, and combined with the technology of artificial intelligence, has become one of the core technologies of today's scientific and technological development. At present, some common famous datasets[44][45][46] contain entities and relations of the order of magnitude of millions of, including information about the location, music, etc., which are widely used in assessment tasks in a number of fields. In some specific assessment tasks, specific datasets are also used, such as the COMBO dataset used in the assessment task of the open KG [47].

The application of KGE in various fields is becoming more and more extensive. The current application of the task in various scenarios have made significant progress. We discuss several types of classic quest scenarios here.

## 3.1 Link Prediction

Link prediction is a core task in KG completion, aimed at predicting missing entity relation pairs (i.e., missing edges) in a KG[48]. Specifically, given a KG G=(V, E), where V is the set of entities and E is the set of relations. The goal is to predict potential new entity relation pairs based on existing entities and relations. In the link prediction task, the functions involved can be divided into three categories: the scoring function, loss function, and evaluation metrics of the model.

### 3.1.1 Scoring function

The scoring function is used to measure the effectiveness of a triplet (head entity, relation, tail entity).

TransE[49]:
$$score(h,r,t) = \|h + r - t\|$$

Consider the relation r as an embedding representing the translation between the head entity h and the tail entity t. If h + r ≈ t, then the score of the triplet (h, r, t) is low, indicating that the triplet is likely correct. On the contrary, a higher score indicates that the triplet is unlikely to be correct.

RotatE [50]:
$$score(h,r,t) = \|h \circ r - t\|$$

Consider the relation r as a rotation operation in the complex plane, and the head entity h should approach the tail entity t after rotating through the relation r. It can handle many types of relations (e.g., symmetric, antisymmetric, inverse, and so on) and is suitable for KGs that handle complex relation patterns.

DistMult[116]:
$$score(h,r,t) = h^T R t$$

Assuming that the relation matrix R is a diagonal matrix simplifies the calculation of the model. Capturing the interaction between entities and relations through matrix multiplication is suitable for handling symmetric relations (such as "friend" relations).

Consider the relation r as a rotation operation in the complex plane, and the head entity h should approach the tail entity t after rotating through the relation r. It is also suitable for handling complex relation patterns.

### 3.1.2 Loss function

The loss function is used to train the model so that positive samples score higher than negative samples.

Contrastive Loss[51]:
$$L = \sum_{(h,r,t)\in\mathcal{G}} \sum_{(h',r',t')\in\mathcal{G}^-} \max(0, \gamma + s(h,r,t) - s(h',r',t'))$$

The contrastive loss aims to learn the representation of samples, so that the representation of positive sample pairs is close and the representation of negative sample pairs is far away. Avoiding overfitting: by introducing a boundary value m, contrastive loss can help the model avoid overfitting and ensure a minimum distance between negative sample pairs. Widely used in metric learning for tasks such as image recognition and natural language processing.

Cross entropy loss[52]:

$$L = -\sum_{i=1}^{N} y_i \log(\hat{y}_i)$$

Cross entropy loss is widely used in classification tasks, especially when the labels are one hot encoding. By minimizing cross entropy loss, the model can learn better parameters to make the predicted probability as close as possible to the true label. Cross entropy loss penalizes the logarithm of probability, allowing the model to focus more on accurate classification with higher probabilities in prediction.

### 3.1.3 Evaluation indicators

Evaluation metrics are used to measure the performance of models in KGE related tasks.

Mean Rank (MR)[53]:

$$\text{MR} = \frac{1}{|\mathcal{T}|} \sum_{(h,r,t) \in \mathcal{T}} \text{rank}(h, r, t)$$

Mean Rank is the average of the correctly predicted rankings. The lower the ranking, the higher the accuracy of the model's predictions. MR does not consider the reciprocal of rankings, making it more sensitive to predictions with lower rankings.

Area Under the Curve (AUC):

$$\text{AUC} = \int_0^1 \text{TPR}(\text{FPR}) d\text{FPR}$$

AUC measures the model's ability to distinguish between positive and negative samples. AUC provides a measure of global performance by calculating the proportion of positive samples scoring higher than negative samples. The closer the AUC value is to 1, the better the model performance.

Mean Average Precision (MAP):

$$\text{MAP} = \frac{1}{|\mathcal{Q}|} \sum_{q \in \mathcal{Q}} \text{AP}(q)$$

MAP is the average precision (AP) of all queries. It measures the overall prediction accuracy of the model in different queries and is particularly suitable for multi label classification tasks.

## 3.2 Entity Alignment

Entity alignment is a key task in KG fusion. Its' goal is to find nodes (entities) representing the same real-world entities in multiple KGs and align them. For example, the "Barack Obama" entity in one KG should correspond to the "Obama president" entity in another KG. Entity alignment is crucial for the integration of KGs, data fusion, and cross-domain graph inference.

### 3.2.1 Knowledge graph embedding model

KGE models map entities and relations in KGs to a low dimensional vector space. The function for this part can refer to the scoring function in III.A, which is affected by the length of the article, so we do not introduce it in detail for the reader here.

### 3.2.2 Similarity based detection function

In entity alignment tasks, it is usually necessary to calculate the similarity of entity embeddings in different KGs to determine whether they represent the same entity.

Cosine Similarity[55]:

$$\text{cosine similarity} = \frac{\mathbf{A} \cdot \mathbf{B}}{\|\mathbf{A}\| \|\mathbf{B}\|}$$

Calculate the cosine value of the angle between two vectors A and B, the larger the value, the higher the similarity. Used to compare the similarity of embedding vectors between two entities.

Euclidean Distance[56] :

$$d(\mathbf{A}, \mathbf{B}) = \sqrt{\sum_{i=1}^{n}(\mathbf{A_i} - \mathbf{B_i})^2}$$

Calculate the distance between two vectors A and B in space, with smaller values indicating higher similarity. Used to measure the distance between entity embeddings.

Manhattan Distance[57]:

$$d(\mathbf{A}, \mathbf{B}) = \sum_{i=1}^{n} |\mathbf{A_i} - \mathbf{B_i}|$$

Calculate the absolute distance between two vectors A and B in space, with smaller values indicating higher similarity. Used to measure the distance between entity embeddings.

### 3.2.3 Evaluation indicator function

In entity alignment tasks, the evaluation indicator functions are commonly used to measure the performance of the model.

Accuracy:

$$\text{accuracy} = \frac{TP + TN}{TP + TN + FP + FN}$$

The proportion of correctly predicted entity alignment quantities to the total entity alignment quantity. Measure the overall performance of the model in entity alignment tasks.

Mean Reciprocal Rank(MRR):

$$\text{MRR} = \frac{1}{Q}\sum_{i=1}^{Q}\frac{1}{\text{rank}_i}$$

For each query entity, calculate the reciprocal ranking of its true matching entities and take the average of all query entities. Measure the performance of the model in many to many entity alignment tasks.

Hits@k:

$$\text{Hits@k} = \frac{C}{Q}$$

Calculate the proportion of correctly aligned entities in the top k results recommended by the model. Measure the recall capability of the model in entity alignment tasks. The KGE model function maps entities and relations in the KG to a low dimensional vector space, providing a foundation for entity alignment. The similarity based detection function can calculate the similarity of entity embeddings in different KGs and determine whether they represent the same entity. The evaluation index function is used to measure the performance of entity alignment models and help select the optimal model.

## 3.3 KG Canonicalization

KG canonicalization is an important task in knowledge storage, sharing, retrieval and application. Its goal is to map multiple entities in KGs to a standardized embedding representation, eliminating redundancy and ambiguity, thereby improving the quality and consistency of KGE. For example, map different emdedding representations such as "New York City", "NYC", and "Big Apple" to a unified embedding representation. KG canonicalization typically involves two main steps. The first step is to learn accurate embedding representations; In the second step, cluster based on the similarity of the embedding representations learned in the first step, and assign a unified embedding representation to each cluster.

### 3.3.1 Knowledge graph embedding model

The KGE model maps entities and relations to a low dimensional vector space, facilitating similarity calculation and clustering. Common embedding models include:

Simplified Embedding Model (SimplE):

$$\text{score}(h,r,t) = \vec{e}_h^T \vec{r} \vec{e}_t + \vec{e}_t^T \vec{r}^{rev} \vec{e}_h$$

SimplE is a model that combines bidirectional information of head and tail entities, capable of handling both symmetric and asymmetric relations. Capturing the interaction between entities and relations through bidirectional embedding is suitable for KGE and entity alignment tasks.

Convolutional Embedding Model(ConvE)[59]:

$$\text{score}(h,r,t) = \mathbf{e}_t^T \cdot \text{vec}(\text{ReLU}(\text{Conv2D}(\mathbf{E}_{hr})))$$

ConvE captures local features of entities and relations through convolutional neural networks, making it suitable for handling complex relation patterns. Utilizing convolution operations to enhance the expressive power of embeddings is suitable for embedding and entity alignment tasks in large-scale KGs.

Quaternion Embedding Model(QuatE)[60]:

$$\text{score}(h,r,t) = \|\mathbf{q}_h \otimes \mathbf{q}_r - \mathbf{q}_t\|_2$$

QuatE captures the rotation and scaling characteristics of entities and relations through quaternion multiplication, making it suitable for handling multimodal relations. Utilizing the high-dimensional representation ability of quaternions to enhance the expression ability of embeddings, it is suitable for KGE tasks with complex relation types.

### 3.3.2 Clustering function

In the process of KG canonicalization, clustering algorithms are used to group aligned entities and select canonical representations. Common clustering methods include K-means, hierarchical clustering, etc.

K-means clustering[61]:

$$J = \sum_{i=1}^{n} \sum_{k=1}^{K} r_{ik} \cdot ||x_i - \mu_k||^2$$

The purpose of K-means clustering is to minimize the objective function J, which is defined as the sum of squared distances from all data points to the center of their respective clusters. Calculate the Euclidean distance from each data point $x_i$ to its cluster center $mu_k$ through double summation, and only consider the data points assigned to a certain cluster through the indicator variable $r_{ik}$. This optimization process aims to make the data points within the cluster as close to their centroids as possible, while keeping the centroids of different clusters as far away as possible. By iteratively updating cluster centers and reallocating data points, K-means ultimately divides the data into K clusters, achieving locally optimal clustering results.

### 3.3.3 Evaluation indicator function

The evaluation indicators for KG Canonicalization include accuracy, recall, F1 score, etc.

Accuracy (Precision):

$$\text{Precision} = \frac{\text{TP}}{\text{TP} + \text{FP}}$$

Accuracy is used to evaluate the accuracy of a model in predicting positive classes. High accuracy indicates that in the samples predicted as positive, most of them are indeed positive. Accuracy is mainly used to focus on the performance of the model in positive class prediction, especially in cases where there are few positive class samples.

Recall rate[63]:

$$\text{Recall} = \frac{\text{TP}}{\text{TP} + \text{FN}}$$

The recall rate is used to evaluate the integrity of the model in identifying positive samples. A high recall rate indicates that the model can capture actual positive class samples well, reducing the situation of missing positive class samples. The recall rate is mainly used to focus on the performance of the model in identifying positive samples, especially in situations where there are few positive samples or the model needs to identify as many positive samples as possible.

F1 score:

$$\text{F1-Score} = 2 \cdot \frac{\text{P} \cdot \text{R}}{\text{P} + \text{R}}$$

The F1 score is used to evaluate the overall performance of the model in positive class prediction. It considers both accuracy and recall, thus balancing the accuracy and completeness of the model in positive class prediction. The range of F1 scores is from 0 to 1, with higher values indicating better performance of the model. The F1 score is particularly suitable for situations where there is an imbalance between positive and negative samples, as it can avoid situations where the model performs too well on one metric while performing poorly on another.

## 4 Techniques of KGE with LLMs

With the emergence of real-world applications, there are many emerging application scenarios for the current KGE application tasks. We depict the differences between the above categories of KGs and traditional KGs in Figure 1. The specific characteristics of the KGE are different in different KG scenarios. Current traditional KGE methods face limitations in handling complex relations, long-tail entities, and dynamic knowledge updates, and struggle to effectively capture semantic information and contextual associations. LLMs can compensate for the shortcomings of traditional KGE in semantic expression and contextual reasoning through their powerful semantic understanding and generation capabilities, thereby enhancing the accuracy of knowledge representation and reasoning. With the introduction of LLMs, novel approaches combining pre-trained language models have emerged in the KGE field, demonstrating significant advantages and development potential in knowledge completion, relation prediction, and cross-domain applications. We classify the utilization of LLMs based on diverse application scenarios of KGE, and elaborate on this in the following chapter.

### 4.1 Classic Knowledge Graph (CKG)

Classic knowledge graph is a structured semantic knowledge base based on graphs, which stores graph data in the form of a triplet structure. At present, CKG has been widely used in various task categories in various fields. CKG can be formally defined as

$$G = (E, R, U)$$

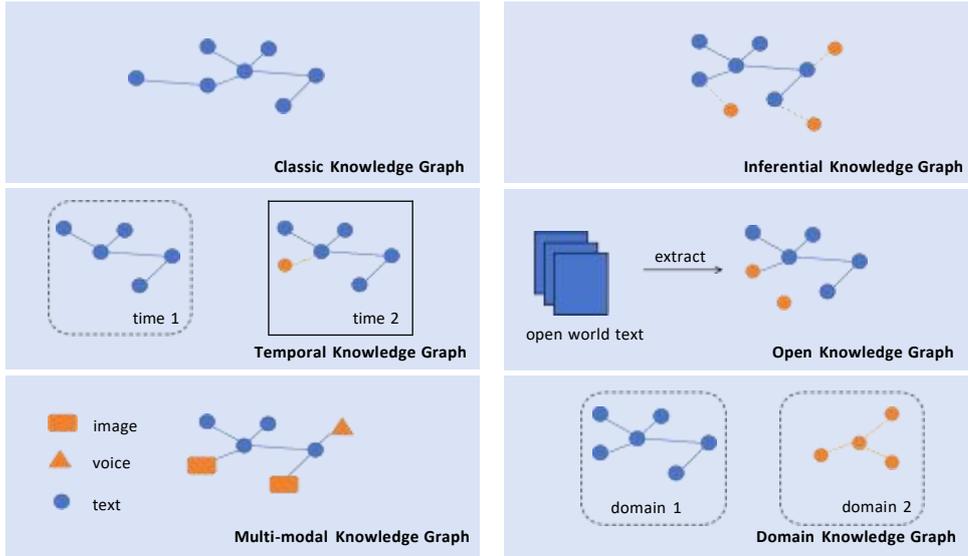

**Fig. 1** KGs in different scenarios

, where E is the set of entities, R is the set of relations and U ⊂ E × R × E represents the set of relation quaternions. With the deepening of research and the emergence of new requirements in the application, a number of KG scenarios that are different from CKG have emerged, which can be introduced one by one in the following sections. Here, in order to distinguish it from KGs in other scenarios, we refer to it as the classic knowledge graph in the following text. Based on the different degrees of use of such methods for LLMs, we can roughly divide KGE methods in CKG scenario into four categories: (1) methods that use LLMs' prompts; (2) methods for fine-tuning LLMs; (3) methods of pre-trained LLMs; (4) methods to use LLMs as agent.

### 4.1.1 Methods that use LLMs' prompts

extract the structural information and any other auxiliary information of the triplet. Then, transform this information into a text description in a suitable format to serve as a prompt input for LLMs. Leverage the text-generation and knowledge-reasoning capabilities of LLMs to facilitate the completion of the task based on KGE. KICGPT [64] uses a contextual learning strategy called knowledge cues, which encodes structured knowledge into presentations to guide LLMs. FireRobBrain [65] includes a prompt module that helps the LLMs generate suggestions for the bot's responses. The combination of LLMs and embeddings of CKGs, facilitated by a well-designed cue module, significantly improves the quality of answers. PEG [66] leverages ensembled KGE through prompts, augments existing methods based on knowledge acquisition, aggregation, and injection, refines evidence through semantic filtering, and then aggregates it into global KGE through implicit or explicit methods. KC-GenRe [67] performs knowledge-enhanced constrained inference in CKG completion tasks, and supports contextual prompts and controlled generation to obtain effective rankings. CP-KGC [68] is designed to adapt prompts to different datasets to enhance semantic richness, and a contextual constraint strategy is adopted to effectively identify multi-aggregated entities in the dataset.

This type of method can effectively utilize the powerful text generation and inference capabilities of LLMs, integrating structured triplet knowledge embeddings from CKGs in a more natural way, thereby providing assistance for complex tasks such as knowledge completion, and potentially uncovering hidden semantic relations between triplets. However, there may be information loss or semantic bias during the process of converting triplets into textual descriptions. Because the triplet structure of CKGs is precise, the accuracy of the original structural information may decrease after conversion to text due to factors such as linguistic ambiguity.

### 4.1.2 Methods for fine-tuning LLMs

collect the CKG triplet data and transform it into a format highly compatible with model training requirements. Then, feed the pre-processed triplet textdata into LLMs. Define suitable loss functions and fine-tune the model parameters in strict accordance with the specific task objectives to optimize the model's performance for accurate and efficient task execution. In order to improve performance while reducing instruction data, DIFT [69] uses a truncated sampling method to select useful facts for fine-tuning and injects KG embeddings into the LLM. ReclIn-Dial fine-tunes [70] large-scale, pre-trained LLMs to generate fluent and varied responses, and introduces knowledge-aware biases learned from entity-oriented CKGs to improve recommendation performance. KG-LLM [71] converts structured CKG data into natural language, and then uses these natural language prompts to fine-tune the LLM to enhance multi-hop link prediction based on embeddings. RPLLM [72] uses the node names of the CKG to fine-tune the relational prediction task of LLMs, enabling the model to operate adequately in a sensing setting by using only the node names. KG-Adapter [73] is a parameter-level CKG ensemble method based on parameter-effect fine-tuning, which introduces a new adapter structure specially designed for decoder LLMs, which can encode CKGs from the perspective of node center and relation center, and perform joint inference with LLMs. LPKG [74] augments LLMs with data obtained from CKG, which is fine-tuned to accommodate complex KG scenarios.

By targeted fine-tuning, the model can learn the semantic relations and knowledge contained in the triplet structure of CKG, thereby improving its performance in CKG tasks and effectively utilizing the data of the CKG to enhance the professionalism of the model. It should be noted that the fine-tuning process requires a large amount of high-quality annotation data. Collecting and organizing CKG triplet data and converting it into a suitable format is time-consuming and labor-intensive. If the data quality is not high or the quantity is insufficient, it may lead to overfitting of the model, resulting in poor generalization ability in unseen real-world data.

### 4.1.3 Methods of pre-trained LLMs

transform the CKG triplets along with other relevant background information into a text format that is highly suitable for subsequent processing. Then, utilize this converted text data for pre-training. In line with the pre-training tasks specifically designed, the model's parameters are updated by calculating the loss between the predicted outcomes and the true labels. This process enables the model to effectively learn the language patterns and semantic relations embeddings within the CKG. LiNgGuGKD [75] employs a label-oriented instruction tuning to train pre-trained LLMs teachers and introduces a hierarchical adaptive contrasting distillation strategy to adjust the node characteristics between the teacher and student networks in a potential space, effectively transferring the understanding of semantics and complex relations. GLAME [76] uses publicly available pre-trained LLMs to utilize a KG augmentation module to discover relevant knowledge that has changed due to editing and obtain its internal embedding representation in the LLMs. Salnikov et al. [77] explore pre-trained text-to-text LLMs, enriching additional information from CKG to answer factual questions.

During the pre-training process, LLMs can delve into the semantic relations and language patterns in CKG, enhancing the language model's understanding and embedding representation ability of CKG, and enriching LLMs' knowledge reserves. However, converting CKG to text format may introduce noise or information loss, which can affect the effectiveness of pre-training and make the learned prior knowledge less accurate.

### 4.1.4 Methods to use LLMs as agent

using LLMs as an agent allows for interactive exploration of relevant entities and relations on CKGs, and inference based on retrieved knowledge. However, this type of method is currently not widely used and deserves further exploration in the future. ToG [78] iteratively performs a beam search on the KG via an LLM agent, discovers the most promising inference path, and returns the most likely inference results.

When LLMs play the role of an agent, they can take advantage of the large amount of knowledge and reasoning patterns obtained in the pre-training process, and combine these with the structured information in CKG to perform more complex knowledge reasoning.

## 4.2 Temporal Knowledge Graph (TKG)

Temporal knowledge graph is a kind of dynamic KG that contains facts that change over time, and is usually represented by a quaternion of the form $\langle head\ entity, relation, tail\ entity, timestamp \rangle$; a TKG can be formally defined as

$$G = (E, R, T, U)$$

, where $E$ is the set of entities, $R$ is the set of relations, $T$ is the set of timestamps, and $U \subset E \times R \times E \times T$ represents the set of relation quaternions. At present, LLMs mainly use the method based on prompts in the field of TKG for KGE tasks, accompanied by local use based on fine-tuning and other methods.

In the LLMs method based on prompts, the data in the TKG is generally processed first, so that it can be effectively understood and processed by LLMs, especially for the processing of time information. The prompts are usually constructed according to the specific time characteristics and other effective characteristics of the time series task. In TKG's question-answering tasks, GenTKGQA [79] guides LLMs to answer the temporal question through two phases: subgraph retrieval and answer generation. The virtual knowledge index is designed, which is helpful for the open-source LLMs to deeply understand the temporal order and structural dependence between the retrieved facts through instruction adjustment. In the temporal relation prediction task, zrLLM [80] inputs textual descriptions of KG relations into LLMs to generate relation embedding representations, and then introduces them into an embedding-based approach. The representation of LLMs authorization can capture the semantic information in the description of the relation. This makes relations with similar semantics, close in the embedding space, enabling the model to recognize zero-shot relations even when no context observes. CoH [81] enhances the prediction ability of LLM by using higher-order history chains step by step, gradually provides high-order historical information, and fuses it with graph neural network (GNN) results to obtain more accurate prediction results. Different from the previous tasks, GenTKG [82] also uses fine-tuning, using an instruction fine-tuning strategy with few sample parameters to align the task of LLMs, and using a retrieval enhancement generation framework to solve the challenge through a two-stage approach.

The prompt-based method does not require a large amount of training data and complex fine-tuning process in the TKG task, and can directly use the ability of the pre-trained LLMs to guide the model to generate results that meet the requirements through well-designed prompts, which can effectively reduce the cost and time of the task. However, the design and selection of prompts are highly demanding, and if the prompts are not accurate, clear or comprehensive, the model may generate inaccurate or irrelevant results, and different prompts need to be designed for different tasks, and the universality is limited. In contrast, the fine-tuning-based method can optimize the pre-trained model according to specific tasks and datasets in TKG-related tasks, so that the model can better adapt to specific fields and tasks, and improve the performance and accuracy of the model on this task, which is worthy of further exploration in the future.

### 4.3 Multi-modal Knowledge Graph (MMKG)

Multi-modal knowledge graph is a kind of KG that combines multiple data modalities (e.g., text, image, video, etc.) to construct and represent facts. It is usually represented in the form of a ternary group, i.e., <head entity, relation, tail entity>, where the entity or relation information is not necessarily single-modal data, but can be multi-modal information describing the entity or relation. Formalized definitions are consistent with traditional KG. MMKG enhances the diversity and accuracy of knowledge representation by enriching modal data. At present, prompt-based LLMs methods are mainly used in the field of MMKG. The features of different modal information are extracted and encoded into the prompts, and then the prompts are input into the LLMs to fuse and reason the multi-modal information according to the prompts to complete the corresponding tasks. In the KG completion task, MPIKGC [83] prompts the LLMs to generate auxiliary text to improve the performance of the model. MML [84] initiates the process by meticulously constructing MMKGs with the specific aim of extracting regional embedding representations that inherently encapsulate spatial heterogeneity features. Subsequently, it deftly employs LLMs to generate diverse multi-modal features pertaining to the ecological civilization model. KPD [85] designs a prompt template with three contextual examples to understand all aspects of a web page for phishing detection. In order to construct a unified KG, Docs2KG [86] extends these nodes with multi-hop queries to retrieve relevant information, thereby enhancing the response prompts to queries.

The prompt-based LLMs method can enhance the understanding and fusion ability of multi-modal information, better handle complex multi-modal tasks, and improve the application effect and value. However, the results maybe inaccurate and unreliable due to the limitations of LLMs themselves, such as hallucinations and insufficient attention to modal information such as images. Suggested LLMs based on pre-training or agents can be explored in the next step.

## 4.4 Inductive Knowledge Graph (IKG)

Inductive knowledge graph includes a source KG and a target KG with specific conditions on relations and entities. The goal of embedding representation learning for inductive KGs is to capture structural patterns from the source KG and transfer them to the target KG. The formal definition is the same as the CKG. At present, in the direction of IKG, LLMs are almost always applied to related tasks in the form of prompts. According to the structure and task design prompts of IKG, the relevant text and prompt words can be input into the LLMs. LLMs can infer and analyze KG according to the prompts and generate the results. Xing et al. [87] construct a fusion inductive framework to effectively alleviate the illusory problem of LLMs, combined with predefined prompts, to provide powerful and intelligent decision support for practical applications. LLAMA3 [88] uses the current generation of LLMs for link prediction, and realizes the new entity link prediction in the IKG by generating prompts, answer processing, entity candidate search and final link prediction. Different from the previous two attempts in the inductive link prediction task, LLM-KERec [89] innovates in the IKG of the industrial recommendation system and designs prompts covering multiple aspects. By introducing an entity extractor, it extracts unified conceptual terms from project and user information. To provide cost-effective and reliable prior knowledge, entity pairs are generated based on the popularity of the entity and specific strategies.

The prompt-based approach can enhance IKG's inductive ability to better handle complex textual information and reasoning tasks with the help of the language understanding and generation capabilities of LLMs. At the same time, hallucinations may occur, resulting in inaccurate or unreliable reasoning results, affecting the quality and application effect of IKGs. More attempts with fine-tuning-based or pre-training-based approaches are recommendation in the future.

## 4.5 Open Knowledge Graph (OKG)

Open knowledge graph is a kind of KG derived from open datasets, user contributions or internet information, emphasizing the openness and scalability of knowledge. The formalized definition is consistent with traditional KG. OKG lowers the threshold of knowledge access through openness and sharing, while facing the problems of data quality and reliability. With the advent of the era of big data, massive data sources continue to be generated. OKGs have gained attention in more and more scenarios, such as responding to cyber threat attacks [90], mental health observation [91], and understanding geographic relations [92]. Based on the different degrees of use of such methods for LLMs, we can roughly divide them into two categories: (1) methods that use LLMs' prompts; (2) methods for fine-tuning LLMs.

### 4.5.1 Methods that use LLMs' prompts

design specific prompts to transform the data in OKG into inputs that meet the requirements. Then, it enables LLMs to generate textual embedding representations, which are then encoded or converted into vector-form embeddings to support subsequent tasks such as knowledge reasoning. Arsenyan et al. [93] provide guided prompt design to utilize LLMs, utilizing electronic medical record notes to construct OKGs. The entities used in the construction of KG are diseases, factors, treatments, and manifestations of coexistence with the patient while experiencing the disease. Xu et al. [94] use LLMs and refinement by medical experts to construct a heart failure KG. Apply prompt engineering to the three phases of pattern design: pattern design, information extraction, and knowledge completion. The best performance can be achieved by designing task-specific prompt templates and two-way chat methods. The method proposed by AlMahri et al. [95] automates the extraction of supply chain information from a variety of public sources and constructs KGs to capture complex interdependencies between supply chain entities. Zero-shot prompts are used for named entity recognition and relation extraction tasks, eliminating the need for extensive domain-specific training. Datta et al. [96] introduce a zero-shot prompt-based approach that uses OpenAI's GPT-3.5 model to extract hyperrelational knowledge from text. Each prompt contains a natural language query that is designed to elicit a specific response from a pre-trained language model.

Prompts can guide LLMs to focus on specific knowledge content and reduce redundant expressions. By accurately designing prompts, LLMs can directly cut into key information and avoid the output of irrelevant knowledge, so as to effectively filter out redundant information that may exist in OKG. However, due to the ambiguity of the expression of OKG itself, prompts may exacerbate the misunderstanding. If the prompts do not accurately fit the ambiguous knowledge statement, LLMs may generate results based on incorrect understanding, amplifying the adverse effects of the ambiguous parts of OKG.

### 4.5.2 Methods for fine-tuning LLMs

the knowledge in OKG is extracted, converted into a format suitable for LLMs input, such as text sequences, and then fine-tuned for training. Fine-tuning is used to make

**Table 1** Symbols and Their Specific Meanings

| Symbol | Specific Meaning |
|---|---|
| h | Head Entity |
| r | Relationship |
| t | Tail Entity |
| $\mathbb{G}$ | Positive sample set |
| $\mathbb{G}^-$ | Negative sample set |
| $\gamma$ | Interval Parameters (margin) |
| $s(h,r,t)$ | The score of triplet $(h,r,t)$ |
| $y_i$ | True Label |
| $\hat{y}_i$ | Predicted Tags |
| T | Test Set |
| $rank(h,r,t)$ | The ranking of triplet $(h,r,t)$ |
| TPR | True Positive Rate |
| FPR | False Positive Rate |
| Q | Query Collection |
| AP(q) | Average Precision of Query q |
| **h**, **r**, **t** | Vector representations of head entities, relationships, and tail entities |
| ∘ | Element-wise multiplication (Hadamard product) |
| Re(·) | Take the real part of a complex number |
| cosine similarity | Cosine similarity |
| $d(\mathbf{A}, \mathbf{B})$ | The distance between vectors **A** and **B** |
| accuracy | Accuracy |
| MRR | Mean Reciprocal Rank |
| Hits@k | Hit rate of the first k hits |
| $\vec{e}_h, \vec{e}_t$ | Embedding vectors of head and tail entities |
| $\vec{r}, \vec{r}^{rev}$ | Embedding vectors of relationship and inverse relationship |
| Conv2D | 2D Convolution Operations |
| ReLU | Rectified Linear Unit Activation Function |
| $\mathbf{q}_h, \mathbf{q}_r, \mathbf{q}_t$ | Quaternion representation of head entity, relationship, and tail entity |
| $\otimes$ | Quaternion Multiplication |
| J | Objective Function (Loss Function) |
| $C_i$ | The i th cluster |
| $\mu_i$ | The center of the i th cluster |
| $D_{single}$ | Single Linkage Distance |
| $D_{complete}$ | Complete Linkage Distance |
| $D_{average}$ | Average Linkage Distance |
| J | Objective function, representing the sum of squared distances from all data points to their respective cluster centers. |
| n | Total number of data points in the dataset. |
| K | Number of clusters. |
| $x_i$ | The i th data point (i = 1, 2,..., n). |
| $\mu_k$ | The center of the k th cluster (centroid). |
| $r_{ik}$ | Indicator variable indicating whether data point $x_i$ belongs to cluster k. If $x_i$ belongs to cluster k, $r_{ik}$ = 1; otherwise, $r_{ik}$ = 0. |
| P | The proportion of sa1m7ples predicted as positive by the model that are actually positive. |
| R | The proportion of samples that are actually positive classes that are correctly predicted by the model as positive classes. |

LLMs learn the embedding representation of entities, relations and other knowledge in OKG, which is used for downstream tasks such as knowledge completion and question answering. UrbanKGent [97] first designs specific instructions to help LLMs understand the complex semantic relations in urban data, then calls geocomputing tools and iterative optimization algorithms to further improve the LLMs' understanding and reasoning ability of geospatial relations, and finally further trains LLMs based on fine-tuning to better complete different types of OKG tasks. LLM-TIKG [98] uses chatGPT's few-shot learning capabilities to achieve data annotation and augmentation, creating a dataset for fine-tuning a smaller language model, using the fine-tuned model to classify the collected reports by subject and extract entities and relations.

Through fine-tuning, LLMs can learn the semantic and structural information in OKG, so as to express knowledge more accurately when generating text, reduce the expression redundancy caused by inaccurate understanding of knowledge. However, if there is bias or noise in the data used in the fine-tuning process, it may lead to the LLMs overfitting these inaccurate information, and then produce more ambiguity or misunderstanding when processing the OKG.

## 4.6 Domain Knowledge Graph (DKG)

Domain knowledge graph is a kind of KG for a specific domain, which can often record specific knowledge in a specific domain and solve problems in a specific domain in a more targeted manner. At present, DKG has been studied on how to be applied to various domains such as medicine [99][100][101] and electric power [102], and has a wide range of significance for the development of specific fields. Based on the different degrees of use of such methods for LLMs, we can roughly divide them into three categories: (1) methods that use LLMs' prompts; (2) methods for fine-tuning LLMs; (3) methods of pre-trained LLMs.

### 4.6.1 Methods that use LLMs' prompts

prompts can help guide LLMs to focus on domain-specific domain knowledge, and this type of approach often requires a specific prompt template to be designed according to the characteristics of a specific domain. This is the most commonly used LLMs methods in DKG research. In DKG of the medical field, KG-Rank [103] is used to enhance medical question answering. This is the first medical question answering enhanced LLMs framework that combines KGE with ranking technology to generate and grow answers, applying a medical prompt to a given question to identify relevant medical entities. In the field of electric power, Wang et al. [104] use LLMs to generate a question-answer dataset specific to the electric power domain. Not only is the relation of the KG triplet incorporated into the question prompt to improve the quality of the model's answers, but the extended question set from the triplet is used as knowledge augmentation data to fine-tune the LLMs. However, these methods only take into account a single domain scenario and often have limitations. Compared with single-domain DKG applications, KGPA [105] takes into account cross-domain KGE scenarios, uses KGs to generate original prompts and adversarial prompts, and evaluates the robustness of LLMs through these prompts.

Based on the prompt, it can focus on domain knowledge, and can guide LLMs to focus on specific fields, such as accurately obtaining medical-related content when building a medical KG. By designing different prompts, it can also adapt to a variety of KGE tasks. However, the domain adaptability of such methods is often limited, and for some specific and highly specialized fields, prompts may not be well understood and provided with valid answers due to insufficient LLMs training data.

### 4.6.2 Methods for fine-tuning LLMs

in the case of continuous generation of special domain data, the fine-tuned model can better adapt to the evolution of knowledge, help the dynamic update of DKGs, and maintain the integrity and timeliness of knowledge. In the KG alignment task, GLAM [106] converts the KG into an alternative text representation of tagged question and answer pairs based on the fine-tuning of LLMs. The generative ability of LLMs is used to create datasets, and an effective alternative retrieval-enhanced generation style method is proposed. Li et al. [107] propose a fine-tuning model for fault diagnosis and system maintenance of KGs, and the main work includes the construction of KGs and fine-tuning datasets for fine-tuning and deployment of LLMs in the industrial field. In addition, in terms of auxiliary fault diagnosis, Liu et [108] al. embed aviation assembly KG into LLMs. The model uses the graph structure big data in DKG to prefix the LLMs. DKG for prefix tuning enables LLMs to be reconfigured online, which avoids a significant amount of computational load.

Through fine-tuning, the large model can better adapt to the KG tasks in a specific domain, and use the annotated data in the domain to further learn and optimize, so as to improve the accuracy, recall and other performance indicators of the tasks in the field. However, there may be a risk of overfitting, if the amount of domain data used in fine-tuning is insufficient or over-trained, the large model may overfit this limited data, resulting in poor performance in unseen domain data or real-world application scenarios, reduced generalization ability, and inability to effectively process emerging situations or data.

### 4.6.3 Methods for pre-training LLMs

the integration of DKG's structured knowledge in the pre-training process can enable the LLMs to directly learn the knowledge contained in the KG, enhance the understanding and mastery of knowledge in specific domains, and provide a better knowledge base for DKG tasks. In the field of automotive electrical systems, in order to provide timely access to structured information, Wawrzik et al. [109] use an electronic dataset from a Wikipedia article for pre-training and optimized with several prompting methods and features. In the field of food testing, FoodGPT [110] introduces an incremental pre-training step that injects knowledge embedding into LLMs. A method for dealing with structured knowledge and scanned documents is proposed. In order to overcome the problem of machine hallucination, a DKG is constructed as an external knowledge base to support retrieval in LLMs.

The pre-trained LLMs method can improve the efficiency and quality of knowledge extraction in DKG tasks, and it can directly extract knowledge such as entities and relations from specific domain texts with strong language understanding and text generation capabilities, and achieve specific specialized extraction in specific domains. However, there may be knowledge illusions and errors, which can generate knowledge that does not match the facts or does not exist, resulting in the existence of false information in DKGs.

# 5 Prospective Directions

At present, in various KGE application scenarios, LLMs have already provided effective solutions to specific problems through different degrees of invocation. There are still many prospective research directions for the application of LLMs in various KGE scenarios in the future, which we will introduce in this chapter for researchers' reference. We sincerely hope that the proposed potential directions can pique the research interests of beginners in this field. In the process of exploration, we have also found some examples in specific processes, which we will incorporate in the discussion of this chapter to facilitate the readers' understanding.

## 5.1 Cross-domain CKG fusion

after investigation, it is found that the current applications of LLMs in the CKG domain are mainly focused on single-domain domain KGs or generalized KGs, and seldom involve cross-domain KG content fusion. For example, leveraging KGs and ranking techniques to enhance medical question answering in the specialized field of medicine [99], or empowering LLMs as political experts with KGs [66]. In future research, the comprehension and reasoning capabilities of LLMs can be utilized to align KG concepts of two domains based on context and background knowledge, e.g., to identify different nodes representing the same entity in two KGs.

## 5.2 Few-shot Learning for TKG

Currently, the application aspects of LLMs for TKG are mainly focused on link prediction tasks or knowledge quiz tasks with sufficient samples, and very few articles apply LLMs to few-shot scenarios for TKG. Current research proposes [111] a novel few-shot solution using KG-driven incremental broad learning, and leverages [98] ChatGPT's few-shot learning capabilities to achieve data annotation and augmentation. The few existing methods attempt to augment the idea of generating rich relational descriptions by LLMs, but neglect the use of temporal attributes for TKG. We suggest that in the future, LLMs can be fine-tuned on a small amount of TKG data to allow the model to capture temporal relations more accurately by adding explicit representations of temporal data. With prompt learning, LLMs can quickly adapt to few-shot scenarios and provide reasonable modeling of timing data.

## 5.3 Application of MMKG tasks for more specific domains

the research of LLMs in domain-specific MMKG is still seriously insufficient, and has only been attempted in a few specific domains. For example, [98] leverages large language models and multimodal KGs to detect phishing webpages with or without logos.

**Table 2** Research resources related to knowledge graph based on big language model

| Reference | Code Resource |
|---|---|
| RecInDial: A Unified Framework for Conversational Recommendation with Pretrained Language Models [70] | https://github.com/Lingzhi-WANG/PLM-BasedCRS |
| Knowledge Graph Large Language Model (KG-LLM) for Link Prediction [71] | https://anonymous.4open.science/r/KG-LLM-FED0 |
| THINK-ON-GRAPH: DEEP AND RESPONSIBLE REASONING OF LARGE LANGUAGE MODEL ON KNOWLEDGE GRAPH [78] | https://github.com/IDEA-FinAI/ToG |
| Unifying Local and Global Knowledge: Empowering Large Language Models as Political Experts with Knowledge Graphs [66] | https://github.com/xymou/PEG |
| KC-GenRe: A Knowledge-constrained Generative Re-ranking Method Based on Large Language Models for Knowledge Graph Completion [67] | https://github.com/wylResearch/KC-GenRe |
| KG-Rank: Enhancing Large Language Models for Medical QA with Knowledge Graphs and Ranking Techniques [99] | https://anonymous.4open.science/r/KGQA-270F |
| From Large Language Models to Knowledge Graphs for Biomarker Discovery in Cancer [100] | https://github.com/rezacsedu/llmstokgsforcancer |
| Multi-perspective Improvement of Knowledge Graph Completion with Large Language Models [83] | https://github.com/quqxui/MPIKGC |
| UrbanKGent: A Unified Large Language Model Agent Framework for Urban Knowledge Graph Construction [92] | https://github.com/usail-hkust/UrbanKGent |
| Docs2KG: Unified Knowledge Graph Construction from Heterogeneous Documents Assisted by Large Language Models [86] | https://docs2kg.ai4wa.com |
| zrLLM: Zero-Shot Relational Learning on Temporal Knowledge Graphs with Large Language Models [80] | https://github.com/ZifengDing/zrLLM |
| GenTKG: Generative Forecasting on Temporal Knowledge Graph with Large Language Models [82] | https://github.com/mayhugotong/GenTKG |

It lacks a targeted design for more domain-specific KGs involving multi-modal problems. LLMs can be useful for cross-modal information fusion to obtain information from multiple perspectives, providing richer and multidimensional semantic understanding. We propose that in future research, LLMs can be used to reason and fuse more domain-specific MMKG tasks based on the relations between text and other modalities (e.g., image, audio, etc.). For example, in the medical field, LLMs are able to combine the analysis results of CT scan images with the text of the patient's medical history to provide more accurate disease diagnosis and personalized treatment plans.

### 5.4 Generic framework design for IKG

Since IKG are relatively new concepts, the current applications of LLMs in IKG scenarios are still extremely underdeveloped, lacking a generic framework design for tasks such as link prediction in this direction. The survey revealed only a limited number of studies based on IKG, such as leveraging current-generation LLMs to achieve new entity link prediction in KGs through prompt generation, answer processing, entity candidate search, and final link prediction [88]. Both KG scenarios involve emergent new knowledge, and LLMs can play an important role in the process of updating new knowledge through their powerful language comprehension and reasoning capabilities. We propose to use LLMs to integrate new knowledge from multiple sources, deal with redundant, conflicting, or missing information, and ensure automatic correction and generation of new knowledge. For example, when confronted with inconsistent descriptions in different texts, the model can determine the most accurate knowledge based on context, timestamp, and other information.

**Table 3** Research resources related to knowledge graph based on big language model

| Datasets | Datasets Resource |
|---|---|
| ICEWS 05-15 [70] | (https://dataverse.harvard.edu/dataset.xhtm?persistentId=doi:10.7910/DVN/28075 |
| GDELT [71] | https://www.gdeltproject.org/data.html |
| YAGO [66] | https://yago-knowledge.org/ |
| UMLS [67] | https://www.nlm.nih.gov/research/umls/index.html |
| MetaQA [99] | https://github.com/yuyuz/MetaQA |
| ReVerb45K [100] | https://github.com/Gorov/REVERB45K |
| FB15k-237 [83] | https://github.com/TimDettmers/ConvE/raw/master/FB15k-237.zip |
| WN18RR [92] | https://github.com/TimDettmers/ConvE/raw/master/WN18RR.zip |

### 5.5 OKG Canonicalization

The redundancy and ambiguity characteristics of OKG can severely limit its related tasks, and the current LLMs are more often applied to the downstream tasks related to OKG, while the role of its expression canonicalization process is neglected. We observe that LLMs have been applied in OKG tasks such as electronic medical record [93], achieving data annotation and augmentation [98], and constructing KGs [94], but their application in the canonicalization tasks of OKG has not yet been identified. We propose to obtain richer knowledge background information based on LLMs, and enhance the soft clustering process for synonymous noun or verb phrase representations based on LLMs, which can help to fully utilize the sufficient pre-training advantages of LLMs.

## 5.6 Unified representation learning framework for multiple types and domains

After investigation, it is found that LLMs in different KG scenarios are often designed based on individual KG task scenarios or task scenarios in individual task domains, but there is a lack of a unified representation learning framework based on LLMs for multiple types and domains. It is recommended that the powerful language comprehension and highly scalable nature of LLMs be fully utilized in future research to establish a universal framework suitable for various task types, improve task generalization, and facilitate the development and integration of various types of KG downstream tasks.

# 6 Datasets and Code Resources

In this section, we meticulously curate a collection of classic datasets applicable to various KGE scenarios, aiming to provide valuable resources for readers. Additionally, we systematically organize the code resources derived from some of the articles cited in this paper, facilitating a more comprehensive understanding and practical application for the audience.

It should be noted that the download links for the following datasets are all presented in Table 3.

## 6.1 Dataset

### 6.1.1 ICEWS 05-15

ICEWS is an integrated conflict early warning system, with data sourced from global news media, government reports, and sociological research. It contains over 5 millions social and political events from 2005 to 2015. ICEWS 05-15 includes about 50k entities (such as individuals, organizations, countries, etc.) and approximately 4 millions event triplets. It records political, social, and economic events on a global scale, suitable for time-series KGs and event prediction tasks.

### 6.1.2 GDELT

GDELT is a global event database developed by Google, with data sourced from global news media reports. It contains over 3 billion event records. GDELT includes about 100k entities (such as countries, cities, organizations, etc.) and approximately 3k millions event triplets. It records events, language, and emotional information on a global scale, suitable for social science research such as conflict analysis and political stability research.

### 6.1.3 YAGO

YAGO is a KG extracted from resources such as Wikipedia, WordNet, and GeoNames. It contains over 1 million entities and over 350k facts. YAGO includes about 1.2 millions entities (such as names, locations, events, etc.) and Approximately 350k triplets. Provide rich semantic information, suitable for tasks such as entity linking, relation extraction, and KG construction.

### 6.1.4 UMLS

UMLS is a comprehensive medical terminology system developed by the National Library of Medicine in the United States. It contains over 300k medical terms and concepts. UMLS includes about 300k medical terms (such as diseases, drugs, treatment methods, etc.) and approximately 1 million triplets. It is used for biomedical text processing, information retrieval, and KG construction.

### 6.1.5 MetaQA

MetaQA is a multi hop question answering dataset for question answering systems, sourced from a movie database. Includes 1-hop, 2-hop, and 3-hop question answering pairs. It contains approximately 4.3k entities (such as movie characters, movie titles, etc.) and about 300k triplets. MetaQA is suitable for machine reading comprehension, question answering system development, and multi hop reasoning tasks.

### 6.1.6 ReVerb45K

ReVerb45K is a set of relation facts automatically extracted from web page text. It contains 4.5k relation instances, about 10k entities and about 45k triplets. ReVerb45K is used for relation extraction and KG construction tasks.

### 6.1.7 FB15k-237

FB15k-237 is a subset of the Freebase KG that has been processed to reduce overlap between the training and testing sets. It contains 15028 entities, 237 relations, and approximately 310k triplets. FB15k-237 is Widely used for KG completion and link prediction tasks.

### 6.1.8 WN18RR

WN18RR is a subset of the WordNet KG that has been revised to increase task difficulty. It contains 40943 entities, 11 types of relations, 40943 entities and approximately 93k triplets. WN18RR is mainly used for research on KG completion and link prediction.

## 6.2 Code Resources

Table I lists some important code resources and open source projects related to the research work discussed in this article. These resources cover multiple cutting-edge research directions combining large-scale pre-trained language models with KG. Through these open-source codes, readers can gain a deeper understanding of the implementation details of related research, reproduce experimental results, or make further extensions and innovations.

# 7 Conclusion

LLMs have been widely used in tasks in multiple scenarios of KGE [111][112][113][114][115]. Currently, this promising field has attracted the attention of more and more researchers. A large number of advanced methods have been proposed. In this paper we propose a new taxonomy for the investigation of the current applications of LLMs in the field of KGE. Specifically, we categorize them according to the KG scenarios in which they are applied and innovate them according to the different degrees of use for LLMs in different scenarios. We not only discuss the use of LLMs in various KGE tasks, but also innovate relevant applications and innovation directions that may be involved in the future. Through this survey, we offer a comprehensive and systematic overview of the application of LLMs within the KGE domain. We sincerely hope that this overview can serve as a valuable guide for future research endeavors.

## Declarations

Ethical approval This declaration is not applicable.

Competing interests I declare that all authors have no competing interests as defined by Springer, or other interests that might be perceived to influence the results and discussion reported in this paper.

Authors' contributions Bingchen Liu wrote the main manuscript text, prepared all figures and tables and provided the methodology. Yuanyuan Fang, Naixing Xu, Shihao Hou and Qian Li provided writing-review and editing. Xin Li provided writing-review and editing and provided funding support.

Funding This research is sponsored by the Key R&D Program of Shandong Province, China (No.2023CXGC010801) and the National Key R&D Program of China (2019YFA0709401).

Availability of data and materials All of the materials including figures is owned by the authors and no permissions are required.

Clinical trial number not applicable.